\documentclass[]{fairmeta}
\usepackage{pifont}
\newcommand{\cmark}{\ding{51}}
\newcommand{\xmark}{\ding{55}}
\newcommand{\pmark}{\ensuremath{\triangle}}
\usepackage{amsmath}
\usepackage{amssymb}
\usepackage{makecell}
\usepackage{atbegshi}
\usepackage{mathtools}
\usepackage{amsthm}
\usepackage{enumitem}
\usepackage{algorithm}
\usepackage{algpseudocode}
\usepackage{booktabs}
\usepackage{array}
\usepackage{tabularx}
\usepackage{wrapfig}
\usepackage{colortbl}
\usepackage{fancyhdr}
\usepackage[most]{tcolorbox}
\usepackage{xcolor}
\usepackage{fvextra}
\usepackage{xurl}
\usepackage{hyperref}
\usepackage{wrapfig}
\usepackage{eso-pic}

\definecolor{BrandPurple}{HTML}{7366CC}
\definecolor{BrandCyan}{HTML}{00B4E5}
\definecolor{LightGray}{HTML}{F5F5F5}
\definecolor{TableRowAlt}{HTML}{EAF4FB}
\definecolor{HeaderGray}{HTML}{888888}

\newcolumntype{L}[1]{>{\raggedright\arraybackslash}m{#1}}
\newcolumntype{C}[1]{>{\centering\arraybackslash}m{#1}}

\newtcolorbox{templatebox}[1]{
    breakable,
    enhanced,
    colback=white,
    colframe=gray!80!black,
    colbacktitle=gray!80!black,
    coltitle=white,
    fonttitle=\bfseries,
    title=#1,
    arc=3mm,
    boxrule=1pt,
    drop fuzzy shadow={gray!50!white},
    left=5mm, right=5mm, top=3mm, bottom=3mm
}

\newtcolorbox{highlight}{
    colback=BrandCyan!5,
    colframe=BrandCyan,
    arc=2pt,
    boxrule=0.8pt,
    left=6pt, right=6pt, top=4pt, bottom=4pt,
    breakable
}

\newtcolorbox{keyfinding}{
    colback=BrandPurple!5,
    colframe=BrandPurple,
    coltitle=white,
    fonttitle=\bfseries,
    title=Key Finding,
    arc=2pt,
    boxrule=1pt,
    left=6pt, right=6pt, top=4pt, bottom=4pt,
    breakable
}

\newtcolorbox{tipbox}[1]{
    colback=amapbg,
    colframe=amapblue,
    coltitle=white,
    fonttitle=\bfseries,
    title=#1,
    arc=2pt,
    boxrule=1pt,
    left=6pt, right=6pt, top=4pt, bottom=4pt,
    breakable
}

\renewcommand{\headrulewidth}{0pt}
\renewcommand{\footrulewidth}{0pt}


\fancypagestyle{firstpage}{%
    \fancyhf{}
    \fancyhead[L]{\small\sffamily\color{amapblue} DreamX Team}
    \fancyhead[R]{\small\color{HeaderGray} June 2026}
    \fancyfoot[C]{\footnotesize\color{gray}\thepage}
    \renewcommand{\headrulewidth}{0.4pt}
    \renewcommand{\footrulewidth}{0pt}
    \renewcommand{\headrule}{\vspace{-4pt}\color{HeaderGray}\hrule width\headwidth height 0.4pt}
}

\newcommand{\method}{DreamX-World 1.0}
\newcommand{\eprope}{E-PRoPE}

\title{%
\method{}: A General-Purpose Interactive World Model
}

\author{DreamX Team}

\abstract{

\method{} is a general-purpose interactive text/image-to-video world model for
controllable long-horizon generation.
It supports camera navigation, revisits to previously observed regions, and
promptable events across photorealistic, game-style, and stylized domains.
Our data engine combines camera-accurate Unreal Engine rendering, action-rich
gameplay recordings, and real-world videos with recovered camera geometry.
For camera control, we introduce \eprope{}, a lightweight variant of projective
positional encoding that retains PRoPE's projective camera geometry while
applying camera-aware attention to spatially reduced tokens.
We convert a bidirectional video generator into a few-step autoregressive world
model using causal forcing, DMD-style distillation, and long-rollout training.
Training on self-generated long-horizon contexts exposes the model to its own
generated history and reduces the style and color drift that accumulates across
autoregressive chunks.
Memory-Conditioned Scene Persistence retrieves earlier views through
camera-geometry-based retrieval, while residual recycling makes the conditioning
path less sensitive to imperfect memory latents.
Event Instruction Tuning adds composable event control, and reinforcement
learning alignment recovers camera control and visual quality after distillation.
With mixed-precision DiT execution, residual reuse, 75\%-pruned VAE decoding,
and asynchronous pipeline parallelism, \method{} reaches up to 16\,FPS on eight
RTX\,5090 GPUs.
On our 5-second basic evaluation, \method{} achieves a camera-control score
of \textbf{73.75} and an overall score of \textbf{84.76}, outperforming
HY-WorldPlay 1.5 and LingBot-World in overall score, which achieve \textbf{80.79}
and \textbf{80.45}, respectively.}

\metadata[GitHub]{\url{https://github.com/AMAP-ML/DreamX-World}}
\metadata[Project Page]{\url{https://dreamx-world.github.io}}

\begin{document}
\maketitle

\thispagestyle{firstpage}

\begin{figure}[b!]
    \centering
    \includegraphics[width=0.92\linewidth]{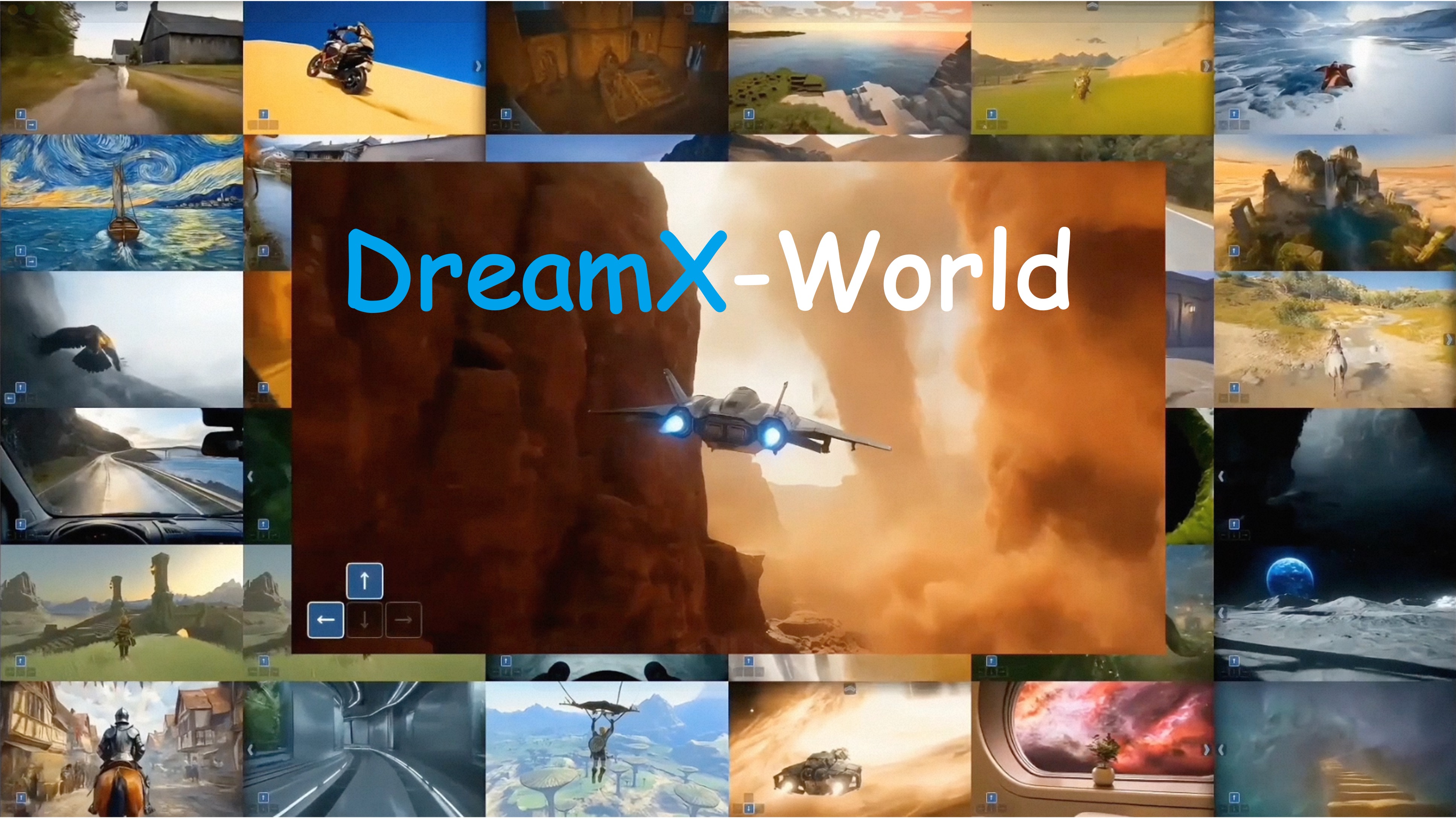}
    \vspace{-0.5em}
    \caption{\textbf{\method{}} generates interactive videos with precise camera and event control
    across photorealistic, game-style, and stylized visual domains.}
    \label{fig:teaser}
    \vspace{-1em}
\end{figure}

\section{Introduction}

World models extend video generation from passive visual synthesis to
interactive simulation. While modern video diffusion models can produce
high-quality short clips, interactive systems must additionally respond to
user controls and preserve scene state over long horizons
\citep{wan2025wan, gao2026lingbotworld, hyworld2025}.
Changing the camera trajectory should reveal a consistent scene rather than
produce an unrelated plausible view; revisiting a location should preserve its
layout and identity; and prompted events should modify the existing world
state. These requirements become particularly difficult when a single model
must operate across photorealistic, game-style, and stylized domains.

Building such a model requires videos spanning different visual domains,
together with reliable camera, action, and event annotations. No single data
source provides this coverage at sufficient scale, motivating the combination
of synthetic, game, and real-world data. Beyond the data problem, three coupled
technical challenges remain.

First, camera control must translate a prescribed trajectory into consistent
viewpoint changes across scenes with different scales and motion distributions
\citep{bahmani2025ac3d}. The conditioning mechanism must provide sufficient
geometric accuracy without substantially increasing the cost of the video
backbone.
Second, interactive generation must preserve scene content beyond the local
context window. Once an earlier view leaves the context, the model may render a
different plausible scene when that region is revisited
\citep{hong2025relic}. Autoregressive generation compounds this problem because
small prediction errors accumulate into appearance, style, and color drift
\citep{huang2025selfforcing}.
Finally, continuous interaction imposes a latency requirement that is absent
from offline video generation. Reducing the number of diffusion steps improves
throughput, but aggressive distillation can weaken visual quality, camera
control, and rollout stability \citep{zhu2026causalforcing}. A practical system
must therefore reduce both sampling and decoding costs while retaining the
capabilities of the bidirectional model.

We present \method{}, a general-purpose interactive world model initialized
from Wan2.2 \citep{wan2025wan}. Its training pipeline progressively
introduces camera conditioning, non-local scene memory, event interaction, and
autoregressive long-video generation, followed by post-distillation alignment.
The data engine combines Unreal Engine trajectories with exact camera geometry,
gameplay recordings with action-rich dynamics, and real-world videos with
recovered camera poses. After geometric filtering and normalization, these
sources provide a common training representation across photorealistic,
game-style, and stylized domains.

\begin{figure}[H]
\centering
\includegraphics[width=0.885\linewidth]{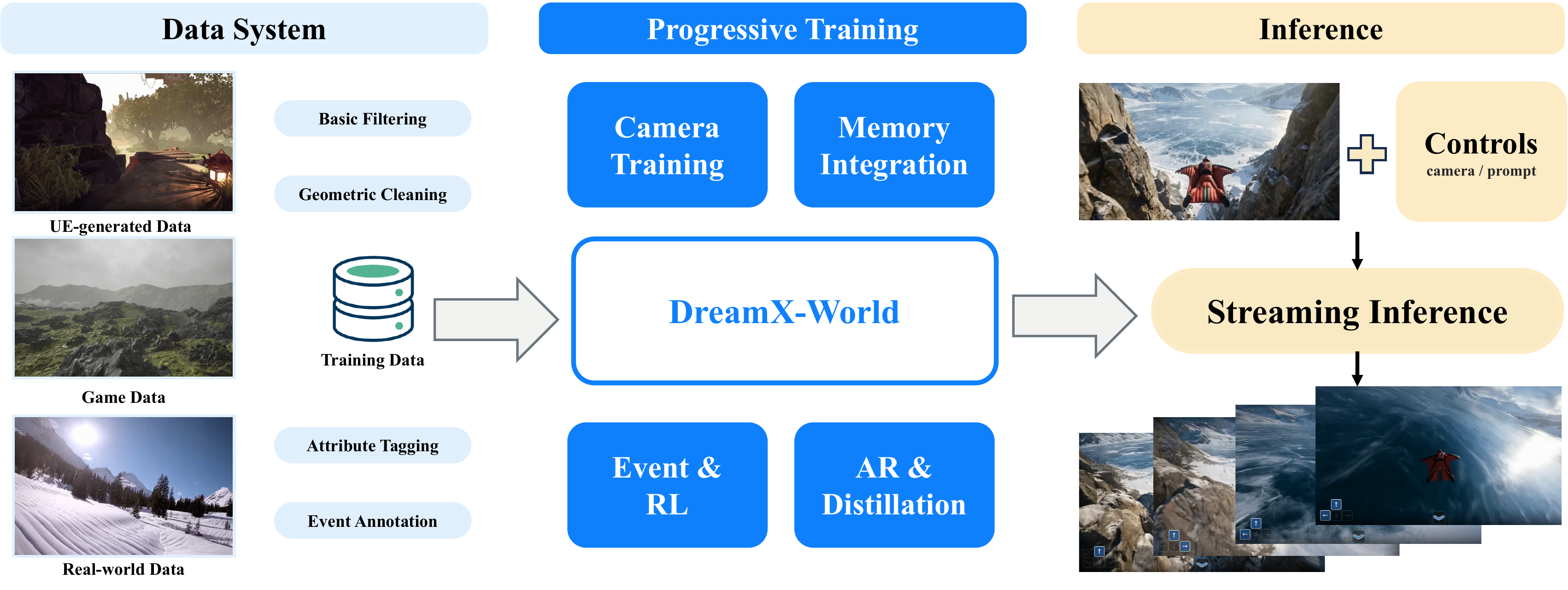}
\caption{\textbf{System overview of \method{}.}
The pipeline integrates camera-accurate data, efficient control, autoregressive
distillation, long-range memory, interaction alignment, and optimized serving.}
\label{fig:pipeline}
\end{figure}

Camera control is introduced through \eprope{}, an efficient variant of PRoPE \citep{li2025cameras}. It computes projective attention on spatially reduced
tokens, retaining comparable trajectory-following performance while reducing
inference latency by approximately 30\%. Once camera motion is controllable, a
memory-conditioned stage addresses a different failure mode: previously
observed content may change after leaving the local context window.
Camera-geometry-based retrieval supplies non-local visual evidence, while
residual recycling improves robustness to imperfect memory latents.

Event Instruction Tuning subsequently adds structured multi-entity event
control through the text-conditioning interface. For streaming generation, the
bidirectional video model is converted into a few-step autoregressive generator
using causal forcing, DMD, and long student rollouts
\citep{zhu2026causalforcing, yin2024onestep, huang2025selfforcing}. This training exposes the model to generated history and reduces the style and color drift that accumulates across chunks.

Because distillation can reduce visual diversity, motion quality, and camera
controllability, the final training stage applies reinforcement learning to the
DMD-distilled model. Camera-control and video-quality rewards guide a
conservative update while preserving the few-step autoregressive interface.
Mixed-precision DiT execution, residual reuse, pruned VAE decoding, and
asynchronous pipeline parallelism then enable streaming generation at up to
16\,FPS on eight RTX\,5090 GPUs.

\begin{highlight}
\textbf{Our main contributions are:}
\begin{itemize}[leftmargin=*, itemsep=1pt, topsep=2pt]
    \item \textbf{Efficient and controllable world generation.}
        We build a multi-source data system and introduce \eprope{} for
        camera-controlled generation. By applying projective conditioning to
        spatially reduced tokens, \eprope{} retains comparable
        trajectory-following performance to PRoPE with approximately 30\%
        lower inference latency.
    \item \textbf{Long-horizon generation with scene persistence.}
        Geometry-guided memory preserves earlier observations during camera
        revisits, while autoregressive distillation and long-rollout training
        reduce the style and color drift accumulated across generated chunks.
        Reinforcement learning further improves the camera control and visual quality
        of the DMD-distilled model without changing its few-step inference interface.
    \item \textbf{Real-time streaming deployment.}
        Together with mixed-precision DiT inference, residual
        reuse, pruned VAE decoding, and asynchronous pipeline parallelism, the
        system reaches up to 16\,FPS on eight RTX\,5090 GPUs.
\end{itemize}
\end{highlight}

\section{Data}
\label{sec:data-system}

\begin{figure}[H]
\centering
\includegraphics[width=0.95\linewidth]{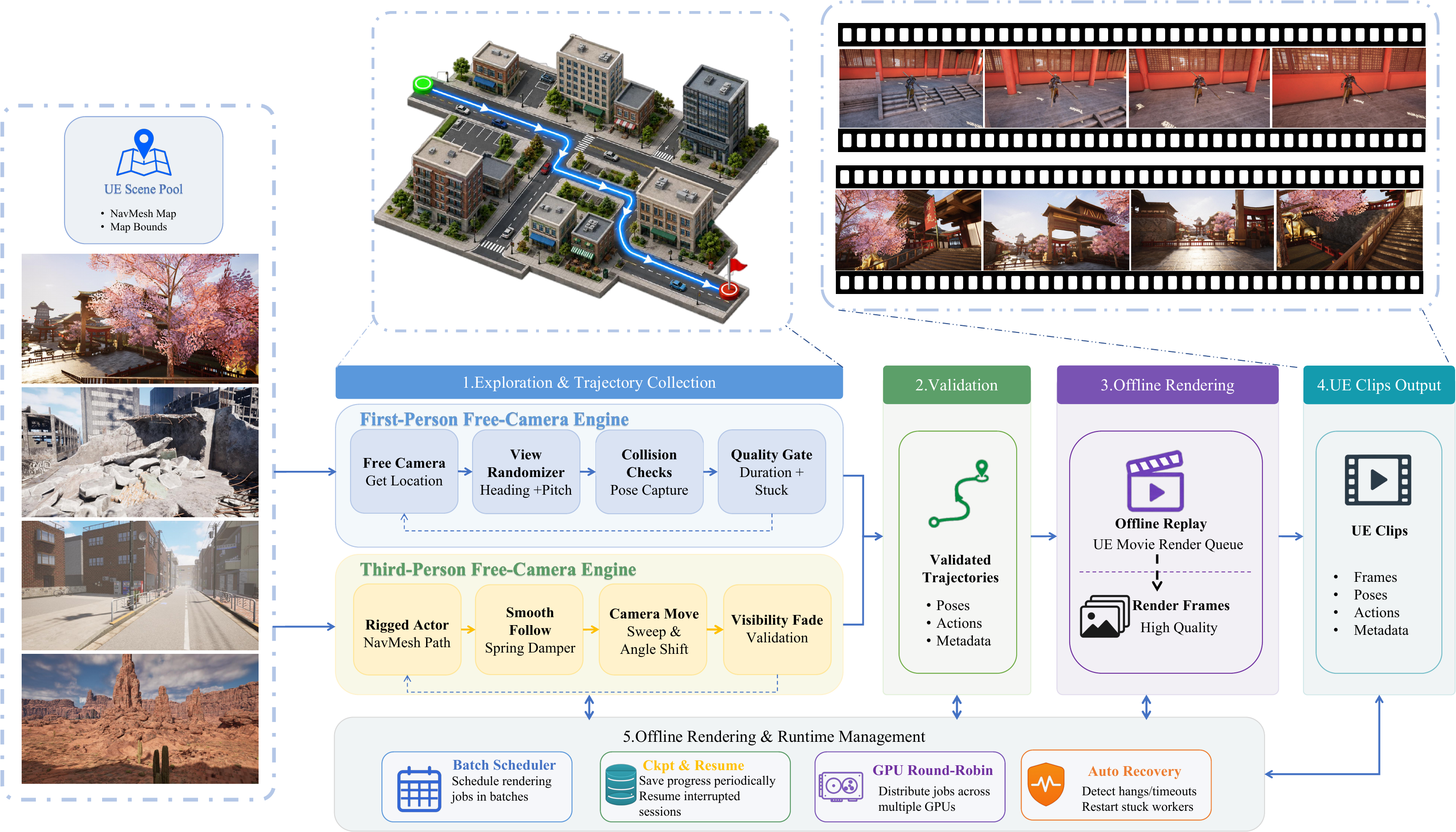}
\caption{UE data generation pipeline. Trajectories are collected and validated online, then rendered offline with camera poses, actions, and metadata. The runtime layer supports distributed rendering and failure recovery.}
\label{fig:ue_pipeline}
\end{figure}

Training interactive world models requires diverse videos with reliable camera and action annotations. Real-world data provides visual diversity but often lacks such annotations, whereas synthetic and game data offer precise control but have limited domain coverage. We therefore combine UE-generated, real-world, and game data in a unified pipeline for annotation, geometric processing, and quality control.

\subsection{Data Curation}

\subsubsection{UE-generated Data}
\label{sec:ue_data}

We synthesize a substantial portion of the training data in UE5 (Unreal Engine 5), which provides controllable camera and agent motion together with ground-truth annotations.

The UE data comprises three observation modes: \textit{first-person}, third-person, and event subsets, which share a unified output schema. The \textit{first-person} subset records free camera exploration, the third-person subset follows a moving character, and the event subset captures object interactions and visible state changes. A distinctive property of the UE subset is its per-frame ground-truth annotation: each frame includes a discrete action vector encoded as keyboard-style control signals (WASD for translation, IJKL for rotation) and camera pose (position and Euler angles). Third-person clips additionally record the character's world position and heading, enabling joint reasoning over camera and agent motion.

UE clips are generated through a two-stage pipeline: first-person and third-person engines explore scenes to discover high-quality trajectories, which are then rendered offline. This decoupled design avoids wasting computational resources on invalid or low-motion clips.

\paragraph{\bf{First-person Free-camera Generation.}}
 
An free camera explores each scene using UE's NavMesh navigation system. It navigates toward sampled goals with randomized heading and pitch adjustments. Collision checking, minimum-duration and path-length constraints, and stuck detection are used to reject invalid trajectories. Accepted trajectories are replayed for offline rendering.

\paragraph{\bf{Third-person Character-driven Generation.}}
A rigged character follows navigation paths while a follow camera records the motion. The camera uses smooth tracking, collision avoidance, and occlusion handling to maintain visibility of the character.

\paragraph{\bf{Rendering and Runtime Management.}}
Validated trajectories are rendered with UE's Movie Render Queue and stored with their poses, actions, and metadata. Rendering jobs are distributed across multiple GPUs with checkpoint resumption and automatic failure recovery.

\subsubsection{Real-world and Game Data}
\label{sec:realworld_pipeline}
\label{sec:game_pipeline}

We collect real-world videos from SpatialVID~\citep{wang2025spatialvid}, RealEstate10K~\citep{zhou2018stereo}, Sekai~\citep{li2025sekai}, and DL3DV~\citep{ling2024dl3dv}. Camera poses are sparsely estimated on several key frames using MegaSaM~\citep{li2025megasam} and interpolated through the pipeline in Section~\ref{sec:cleaning}. Game data is collected from Sekai-Game~\citep{li2025sekai} and OmniWorld-Game~\citep{zhou2025omniworld}. Their engine-exported poses are converted to the same camera coordinate system as the UE and real-world data.

\subsection{Data Annotation and Filtering}
\label{sec:cleaning}

Our data processing pipeline consists of three-stage quality control: basic filtering, geometric camera-pose cleaning, and video captioning and attribute tagging.

\paragraph{\bf{Basic Filtering.}}
We remove clips with insufficient duration or frame rate, excessive overlaid text, black borders, or limited visual change. Visual change is measured using the cosine similarity between CLIP embeddings of the first and last frames.

\begin{wrapfigure}{r}{0.45\textwidth}
\centering
\includegraphics[width=0.4\textwidth]{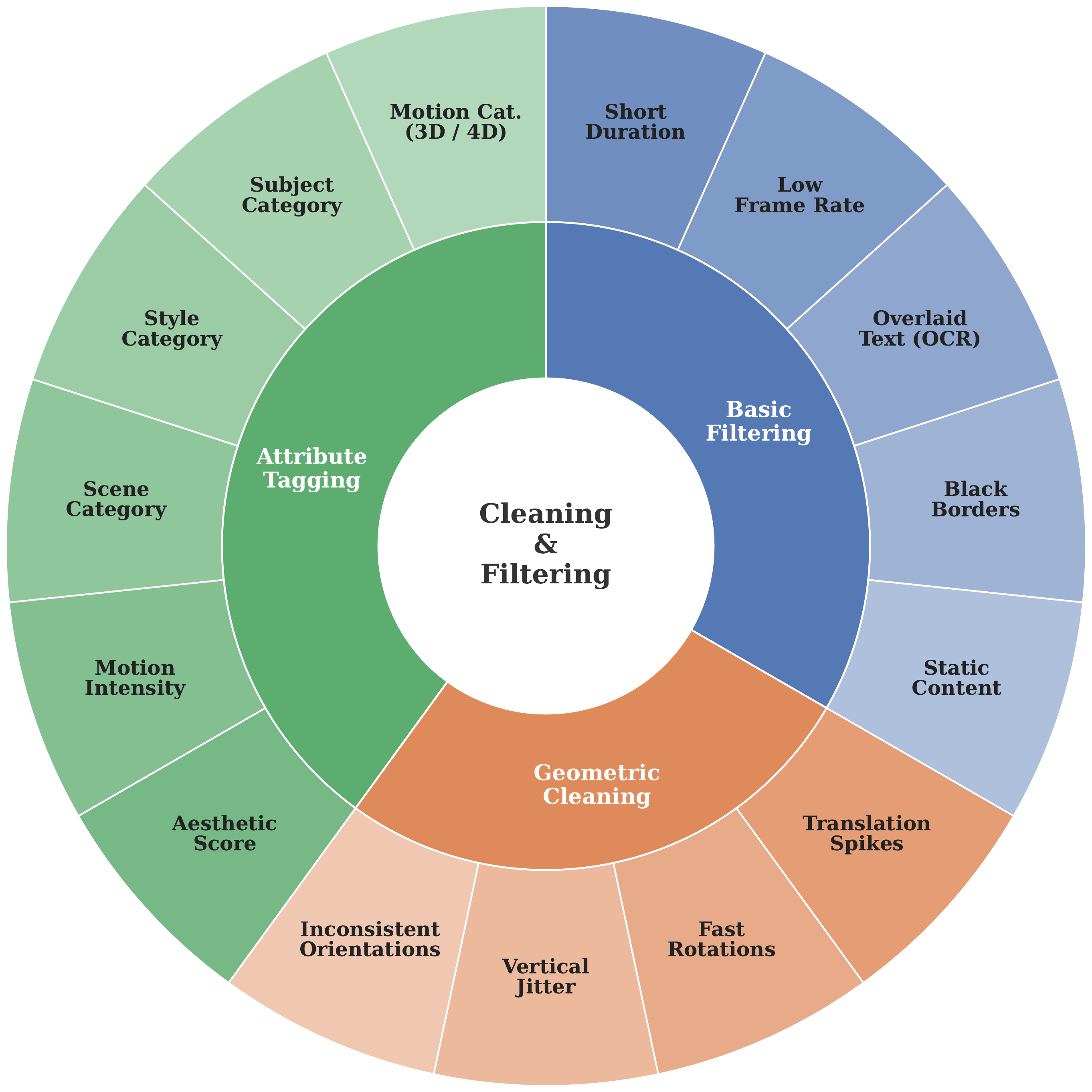}
\caption{Overview of the cleaning, filtering, and attribute tagging pipeline.}
\label{fig:cleaning_pie}
\vspace{-10pt}
\end{wrapfigure}

\paragraph{\bf{Geometric Camera-pose Cleaning.}}
For real-world videos, sparse camera poses are densified using SLERP for rotation and linear interpolation for translation. Each trajectory is normalized and checked for inconsistent intrinsics, translation spikes, rapid rotations, vertical jitter, and invalid orientations.

\paragraph{\bf{Video Captioning and Attribute Tagging.}}
Each clip is annotated with a global caption describing the scene, subjects, actions, and temporal changes. Retained clips are further tagged with aesthetic quality, motion intensity, scene category, visual style, subject type, and motion category. The motion category distinguishes static scenes with camera motion (3D) from scenes containing both camera and object motion (4D). These annotations and tags are used for filtering and training.

\subsection{Event Instruction Data}
\label{sec:event-data}

For Event Instruction Tuning, we select high-quality clips that contain visible state
changes from the cleaned and tagged data pool and annotate them with structured event
descriptions.
Following a hierarchical captioning strategy, each annotation pairs a holistic
\emph{global description} with dense, time-aligned entity-level event records.
The global description summarizes the static scene context and overall temporal evolution,
while each event record specifies the \emph{entity reference}, \emph{event predicate},
\emph{spatial anchor}, and \emph{temporal interval} of a dynamic change.
For composable-event examples, each participating object is assigned its own event record,
and inter-object interactions (e.g., collision, handoff) are explicitly described in the
global caption.
The dataset mixes single-object events and composable events so that the model learns to
ground both atomic and compositional instructions.

\section{Progressive Training Pipeline}

We initialize \method{} from the Wan2.2-TI2V model~\citep{wan2025wan} and progressively adapt it for camera control, memory conditioning, event interaction, and autoregressive generation.

\subsection{Camera-Aware Training}
Since Wan2.2-TI2V does not take camera trajectories as input, we first train
the bidirectional model on pose-annotated videos to support explicit 6-DoF
camera control.
Projective Positional Encoding (PRoPE)~\citep{li2025cameras} introduces a relative conditioning mechanism by introducing both inter-camera frustum relationships and camera-agnostic token positions (e.g., RoPE~\citep{su2024roformer}) directly into transformer self-attention blocks.
Mathematically, for a sequence of input tokens $X=\{x_s\}_{s=1}^{S}$, $X\in R^{S\times d}$, $x_s\in R^{d}$, it defines a per-token matrix $D_s^{PRoPE} \in \mathbb{R}^{d \times d}$  composed of two complementary submatrices of shape $\frac{d}{2} \times \frac{d}{2}$:
$$D_s^{PRoPE} = \begin{bmatrix} D_s^{Proj} & 0 \\ 0 & D_s^{RoPE} \end{bmatrix}.$$
The first submatrix, $D_s^{Proj}$, encodes the full projective camera geometry using the world-to-image projection matrix.
The second submatrix, $D_s^{RoPE}$, replicates the standard rotary position embeddings (RoPE). The matrix $D_s^{PRoPE}$ is applied to the attention query and key tokens via matrix multiplication.

\begin{wrapfigure}[16]{r}{0.45\textwidth}
    \vspace{-1.0\baselineskip}
    \centering
    \includegraphics[width=\linewidth]{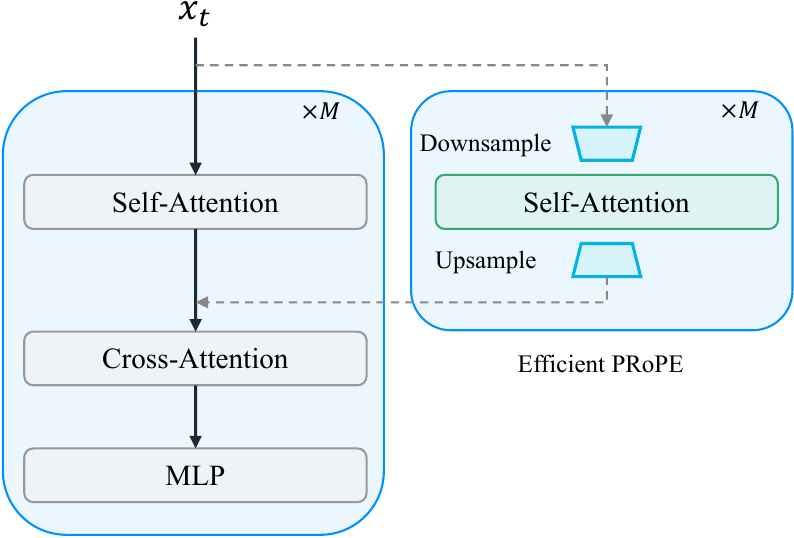}
    \caption{Our Efficient PRoPE (\eprope{}) component is attached to each DiT attention layer. We neglect the adaLN modulation for simplicity.}
    \label{fig:efficient-prope}
\end{wrapfigure}

Directly applying PRoPE can be expensive on long videos for both training and inference, as it requires adding extra attention modules to the DiT backbone in a layer-wise manner, nearly doubling the overall computational cost.
We argue that PRoPE primarily captures the view-dependent high-level semantics.
Therefore, computing attention over the full-resolution video token set, which contains complete semantics, is unnecessary.
To this end, we introduce a lightweight variant of PRoPE as displayed in Figure~\ref{fig:efficient-prope}. The key idea is to focus on a downsampled set of tokens. Specifically, we downsample the PRoPE attention input tokens $X^{PRoPE}$ along the spatial dimension and project them into a lower-dimensional query/key/value space, yielding
$X^{PRoPE} \in \mathbb{R}^{N \times d'}$ where $d' < d$ and $N$ is the number of downsampled tokens.  This makes camera control highly efficient during both training and inference  while retaining most of the controllability. For example, given a 5-second 720P video, the VAE of Wan2.2 5B maps it to $S=18480$ tokens, while we downsample it to $N=4096$ tokens before computing PRoPE attention, indicating a more than 4.5x spatial downsampling ratio. Consequently, it significantly reduces the training and inference time by approximately 50\% and 30\% respectively.

\paragraph{\bf{Efficient PRoPE (\eprope{}).}}

Additionally, we omit the $D_s^{RoPE}$ component and use only the projective submatrix $D_s^{Proj}$, since the original attention in
the DiT backbone already provides sufficient spatiotemporal inductive bias for fine-grained semantic modeling. After PRoPE attention, we upsample the output tokens back to the original resolution and simply add them to the original DiT attention output. With standard denoising objective, we train PRoPE modules by freezing the DiT backbone and only backpropagating the gradients to the PRoPE parameters.
We compare the performance of PRoPE and \eprope{} in Table~\ref{tab:prope}.

Interestingly, during inference, we empirically find that the downstream model can still leverage the pre-trained PRoPE component in a plug-and-play manner even when trained without it, suggesting the strong and robust geometric bias of PRoPE.

\begin{table}[H]
\caption{Comparing PRoPE and \eprope{} on Omni-WorldBench~\citep{wu2026omniworldbench}. \eprope{} achieves comparable camera control performance to the full PRoPE while being more computationally efficient. Latency is measured as the average time (in seconds) to generate a 5-second video at $1280 \times 720$ resolution on 8 NVIDIA H20 GPUs.}
\label{tab:prope}
\centering
\renewcommand{\arraystretch}{1}
\renewcommand{\tabularxcolumn}[1]{m{#1}}                                  
\footnotesize
\setlength{\tabcolsep}{2.8pt}
\begin{tabularx}{\linewidth}{@{}p{.9in}*{7}{>{\centering\arraybackslash}X}@{}}
\toprule
\rowcolor{amapblue}
\textcolor{white}{\textbf{Model}} &
\textcolor{white}{\textbf{\shortstack{Camera\\Control $\uparrow$}}} &
\textcolor{white}{\textbf{\shortstack{Image\\Quality $\uparrow$}}} &
\textcolor{white}{\textbf{\shortstack{Dynamic\\Degree $\uparrow$}}} &
\textcolor{white}{\textbf{\shortstack{Transition\\Detect $\uparrow$}}} &
\textcolor{white}{\textbf{\shortstack{Temporal\\Flicker $\uparrow$}}} &
\textcolor{white}{\textbf{\shortstack{Motion\\Smooth. $\uparrow$}}} &
\textcolor{white}{\textbf{\shortstack{Latency $\downarrow$}}} \\
\midrule
\rowcolor{TableRowAlt}
PRoPE & 73.89 & 66.15 & 87.5 & 96.67 & 96.02 & 98.65 & 80 \\
\eprope{} & 73.75 & 66.75 & 85.83 & 98.33 & 96.17 & 98.79 & 59 \\
\bottomrule
\end{tabularx}
\end{table}

\FloatBarrier
\subsection{Memory-Conditioned Scene Persistence}
\label{sec:memory-training}

\begin{figure}[t]
\centering
\includegraphics[width=0.8\linewidth]{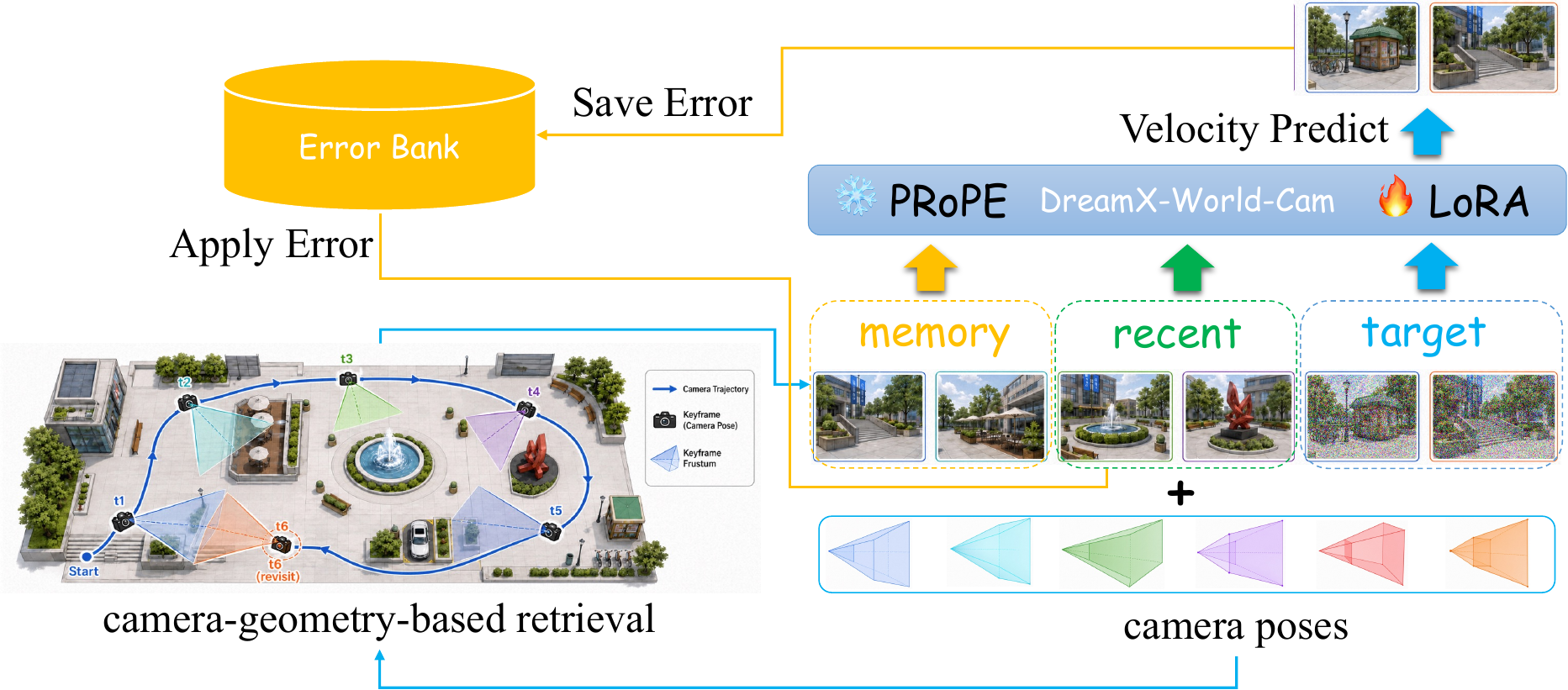}
\caption{Training framework for Memory-Conditioned Scene Persistence.
Geometry-based retrieval selects non-local memory frames.
The memory frames are packed with recent history frames and target frames into
the same DiT self-attention stream.
Supervision is applied only to the target prediction.
The residual-recycling path perturbs conditioning tokens without changing the
supervised target.}
\label{fig:memory-training-framework}
\end{figure}

This training stage targets inconsistency when the model generates the current
frame.
As shown in Fig.~\ref{fig:memory-training-framework}, we train the Diffusion
Transformer (DiT) to use both recent context and
retrieved memory frames.

\subsubsection{Memory Conditioning}

We use two sources of conditioning.
Recent history frames are the latest denoised latent frames before the
target window, while memory frames are predicted clean latent frames retrieved
from earlier history.
The target frames are the noisy latent frames that the model needs to
denoise and supervise.
For a latent video sequence \(z^0_{1:T}\) with camera signals \(\pi_{1:T}\),
we sample target latent frames \(z_\mathcal{C}\), recent history latent frames \(z_\mathcal{H}\),
and memory latent frames \(z_{\mathcal{M}}\), and form the model input sequence
\begin{equation}
    z_{\mathrm{pack}}
    =
    [\,z_{\mathcal{M}}
      \mid z_{\mathcal{H}}
      \mid z^{\tau}_{\mathcal{C}}\,],
    \label{eq:memory-conditioned-input}
\end{equation}
latents, 
\(z^\tau_{\mathcal{C}}\) are target latents with diffusion noise level
\(\tau\), and \([\,\cdot\mid\cdot\,]\) denotes concatenation along the token dimension. During training, we update the model parameters with standard rectified flow training objective and only compute loss on target latent frames.
We also compare other memory injection mechanism, such as cross-attention and VACE-style conditioning \citep{jiang2025vace}. However, these approaches consistently produces worse generation performance. 


\subsubsection{Memory Retrieval}

We retrieve memory frames by geometry-based clues. In specific, we use camera pose and view overlap to choose history frames that are highly relevant to the target view, instead of selecting frames only by temporal distance.
After retrieval, each memory frame is added with RoPE embedding that corresponds to its original temporal location in the generated video when packed into the input sequence.
This prevents distant memory frames from being treated as the immediate frames temporally adjacent to the target frames.
For large time gaps, we use a lightweight temporal-position treatment, inspired
by long-range positional encoding methods such as NTK-aware RoPE scaling, YaRN
\citep{peng2023yarn}, and randomized positional encodings
\citep{ruoss2023randomized}.

\subsubsection{Exposure Bias in Memory}

Memory conditioning has an exposure-bias problem, which originates from the gap between training and inference.
During training, those conditioning frames are sampled from training data, while, during inference, the conditioning frames are generated by model and contain
prediction errors.

We mitigate this train-test gap by adopting the error injection approach proposed in Stable Video Infinity (SVI) \citep{li2025stablevideoinfinity}.
We perturb only the conditioning tokens while keeping the target latent
clean.
At this point, the model learns to draw upon the sampled memory frames when it helps and fall back to its learned prior when the memory frames contain explicit errors.

\subsection{Event Instruction Tuning for Composable Events}

Recent interactive world models expose event-related controls at different
granularities.
LingBot-World~\citep{gao2026lingbotworld} demonstrates promptable global and
local world events, HY-WorldPlay 1.5~\citep{hyworld2025,sun2025worldplay}
supports text-triggered dynamic events during streaming generation, and
Yume-1.5~\citep{mao2025yume15} introduces text-controlled event editing.
Matrix-Game 3.0~\citep{wang2026matrixgame} focuses instead on action-conditioned
long-horizon memory and real-time streaming.
Related spatially controllable video-generation work, such as
Omni-Effects~\citep{mao2025omnieffects}, studies compositional visual effects
specified by category and location, but does not target persistent interactive
world events under navigation.
However, existing public systems do not provide explicit \emph{composable
events}---multiple objects appearing, acting, and interacting within the same
generation under structured event instructions
(Table~\ref{tab:event-comparison}).
Composable events are essential for realistic world simulation, where
meaningful state changes rarely involve a single object in isolation:
a traffic scene requires pedestrians, vehicles, and signals to respond
concurrently, and an indoor scene may involve multiple characters acting and
reacting to one another.
\method{} introduces an Event Instruction Tuning stage that closes this gap:
users describe the coarse region or relation in which each object appears,
what it does, and how objects interact, and the model responds to all specified
events in a single forward pass.

\begin{table}[!htbp]
\caption{Comparison of event-control capabilities across interactive world
models.
Composable events subsume multi-object and multi-event settings: the model
handles multiple objects with distinct actions and mutual interactions in one
generation.}
\label{tab:event-comparison}
\centering
\renewcommand{\arraystretch}{1.2}
\scriptsize
\setlength{\tabcolsep}{2pt}
\begin{tabularx}{\linewidth}{@{}p{0.95in}*{5}{>{\centering\arraybackslash}X}@{}}
\toprule
\rowcolor{amapblue}
\textcolor{white}{\textbf{Model}} &
\textcolor{white}{\textbf{\shortstack{Promptable\\Events}}} &
\textcolor{white}{\textbf{\shortstack{Object-Level\\Events}}} &
\textcolor{white}{\textbf{\shortstack{Region-Guided\\Events}}} &
\textcolor{white}{\textbf{\shortstack{Multi-Entity\\Composition}}} &
\textcolor{white}{\textbf{\shortstack{Inter-Object\\Interaction}}} \\
\midrule
\rowcolor{TableRowAlt}
LingBot-World & \cmark & \cmark & \pmark & \pmark & \xmark \\
HY-WorldPlay 1.5 & \cmark & \cmark & \xmark & \pmark & \xmark \\
\rowcolor{TableRowAlt}
Matrix-Game 3.0 & \xmark & \xmark & \xmark & \xmark & \xmark \\
Yume-1.5 & \cmark & \cmark & \xmark & \xmark & \xmark \\
\rowcolor{TableRowAlt}
\textbf{\method{}} & \cmark & \cmark & \cmark & \cmark & \cmark \\
\bottomrule
\end{tabularx}
\vspace{2pt}
\scriptsize{\pmark indicates qualitative or partial support without structured
per-object event instructions. Object-level events denote single object/category
events; multi-entity composition requires distinct entities and actions in one
structured instruction. Region-guided events refer to coarse regions or relative
placement rather than coordinate-level localization.}
\end{table}

\paragraph{\bf{Training.}}
Using the event instruction data described in Section~\ref{sec:event-data},
we fine-tune the full DiT while keeping the architecture unchanged.
Event semantics enter exclusively through the text-conditioning interface,
where structured event instructions are rendered as natural-language prompts
covering the global scene and per-entity dynamics.
The tuning mixture combines event-instruction samples with non-event training
clips, which preserves the model's general world-generation capability while
adding responsiveness to promptable events.
We use conservative updates and strict gradient clipping to avoid disrupting
the pretrained visual and motion priors.

\subsection{Autoregressive Long Video Generation and Distillation}

\begin{figure}[H]
    \centering
    \includegraphics[width=0.8\linewidth]{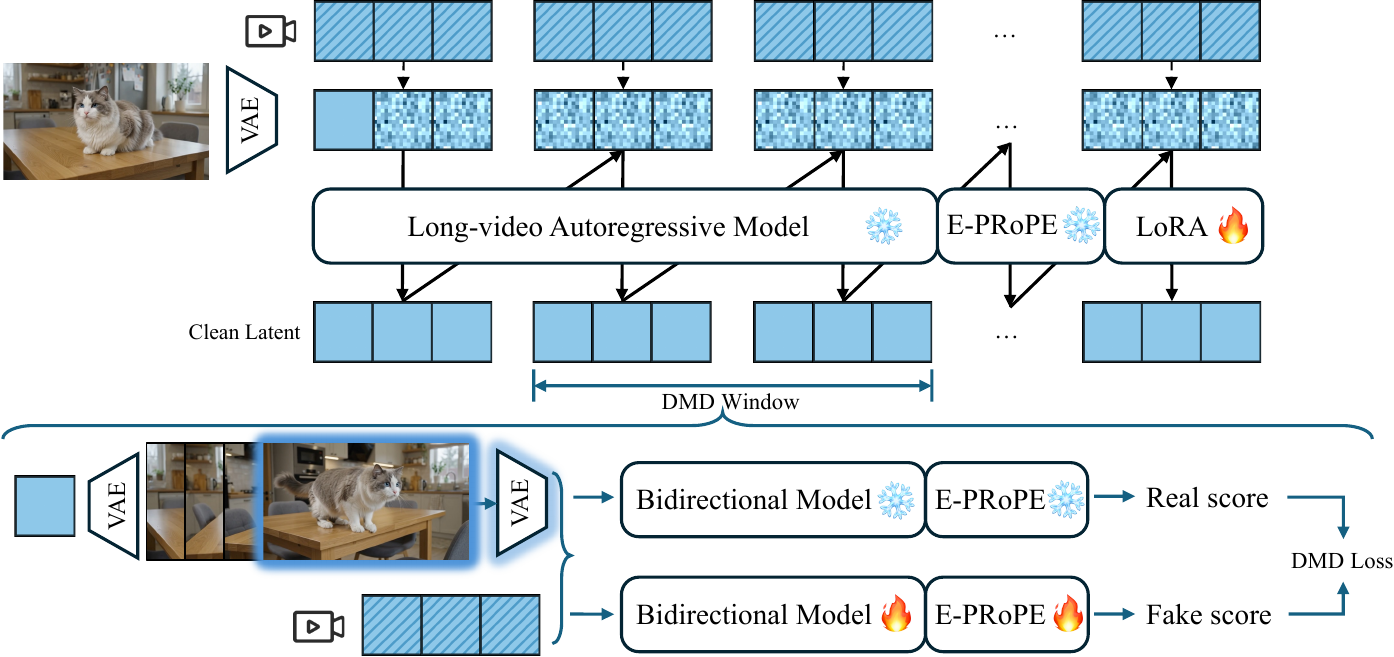}
    \caption{DMD-forcing for camera-controlled long-video distillation. The E-PRoPE AR student is distilled from the bidirectional E-PRoPE teacher through DMD supervision over local temporal windows sampled from long videos, while preserving the streaming autoregressive sampling interface.}
    \label{fig:dmd-forcing}
\end{figure}

To enable few-step long-video generation, we distill a bidirectional model into an autoregressive generator that can stream from generated history while preserving visual quality and camera controllability.

We train few-step autoregressive model using causal forcing~\citep{huang2025selfforcing,zhu2026causalforcing} on large-scale high-quality video data, while keeping it close to the original visual distribution of the bidirectional model. 
Subsequently, following LongLive~\citep{yang2025longlive}, we further adapt the model on long sequences with long rollouts and local temporal windows, using Infinity-RoPE \citep{yesiltepe2025infinityrope} to support extended autoregressive context and reduce long-video failures such as identity drift, background mutation, and weakened prompt or motion control.

For camera-controlled generation, we incorporate an E-PRoPE branch into the long-video T2V student and distill the resulting few-step camera-controlled autoregressive model from a bidirectional E-PRoPE teacher. 
Although this branch enables camera control, chunk-wise inference may still lead to reduced motion smoothness and degraded camera controllability over long sequences. 

We therefore repeat the long-video DMD training to improve this behavior. 
Figure~\ref{fig:dmd-forcing} illustrates the DMD-forcing pipeline, in which camera-controlled student rollouts are matched to the bidirectional E-PRoPE teacher over local temporal windows sampled from long videos.

To preserve I2V quality, we perform I2V DMD distillation using the bidirectional E-PRoPE model as the teacher. The first latent frame of each sampled DMD window is decoded by the VAE and fed to the teacher as the image condition, enabling the teacher to supervise the camera-controlled AR student over local temporal windows of long videos.
With this design, the model can perform stable long-duration inference for videos up to one minute while maintaining camera controllability and temporal coherence across chunks.

\FloatBarrier


\subsection{Reinforcement Learning}

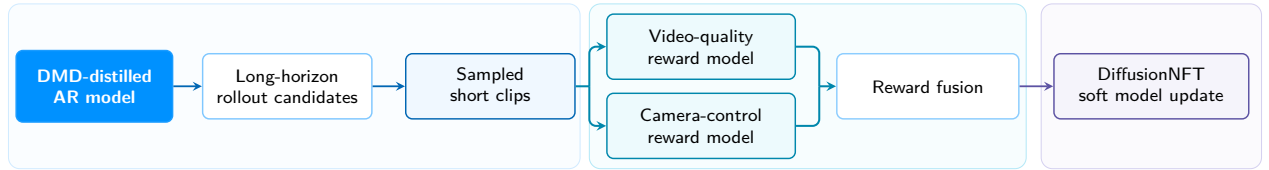
\begin{figure}[!htbp]
\centering
\resizebox{\linewidth}{!}{%
\begin{tikzpicture}[
    font=\sffamily\footnotesize,
    box/.style={
        draw=amapblue!50,
        line width=0.65pt,
        rounded corners=3pt,
        fill=white,
        align=center,
        inner sep=5.5pt,
        minimum height=1.02cm
    },
    model/.style={
        box,
        draw=amapblue!90,
        fill=amapblue,
        text=white,
        font=\sffamily\bfseries\footnotesize,
        minimum height=1.12cm
    },
    reward/.style={
        box,
        draw=BrandCyan!75!black,
        fill=BrandCyan!7
    },
    update/.style={
        box,
        draw=BrandPurple!80!black,
        fill=BrandPurple!7
    },
    clip/.style={
        box,
        draw=amapblue!70!black,
        fill=amapblue!6
    },
    panel/.style={
        draw=amapblue!18,
        line width=0.45pt,
        rounded corners=5pt,
        fill=amapblue!2
    },
    rewardpanel/.style={
        panel,
        draw=BrandCyan!30,
        fill=BrandCyan!3
    },
    updatepanel/.style={
        panel,
        draw=BrandPurple!30,
        fill=BrandPurple!3
    },
    arrow/.style={
        ->,
        >=stealth,
        draw=amapblue!70!black,
        line width=0.8pt
    },
    rewardline/.style={
        draw=BrandCyan!75!black,
        line width=0.9pt,
        rounded corners=3pt
    },
    rewardarrow/.style={
        ->,
        >=stealth,
        draw=BrandCyan!75!black,
        line width=0.9pt,
        rounded corners=3pt
    },
    updatearrow/.style={
        ->,
        >=stealth,
        draw=BrandPurple!80!black,
        line width=0.8pt
    },
    note/.style={
        font=\sffamily\scriptsize\bfseries,
        text=amapfg!72,
        align=center
    }
]

\node[panel, minimum width=8.95cm, minimum height=2.58cm] at (2.83,0) {};
\node[rewardpanel, minimum width=6.84cm, minimum height=2.58cm] at (10.87,0) {};
\node[updatepanel, minimum width=3.45cm, minimum height=2.58cm] at (16.25,0) {};

\node[model, minimum width=2.45cm] (model) at (-0.30,0)
    {DMD-distilled\\AR model};
\node[box, minimum width=2.65cm] (rollouts) at (2.72,0)
    {Long-horizon\\rollout candidates};
\node[clip, minimum width=2.65cm] (clips) at (5.90,0)
    {Sampled\\short clips};
\node[reward, minimum width=2.95cm] (quality) at (9.20,0.62)
    {Video-quality\\reward model};
\node[reward, minimum width=2.95cm] (camera) at (9.20,-0.62)
    {Camera-control\\reward model};
\node[box, minimum width=2.85cm, draw=amapblue!45] (reward) at (12.75,0)
    {Reward fusion};
\node[update, minimum width=3.05cm] (update) at (16.25,0)
    {DiffusionNFT\\soft model update};

\draw[arrow] (model) -- (rollouts);
\draw[arrow] (rollouts) -- (clips);
\coordinate (rewardSplit) at (7.45,0);
\coordinate (rewardMerge) at (11.05,0);
\draw[rewardline] (clips.east) -- (rewardSplit);
\draw[rewardarrow] (rewardSplit) |- (quality.west);
\draw[rewardarrow] (rewardSplit) |- (camera.west);
\draw[rewardline] (quality.east) -| (rewardMerge);
\draw[rewardline] (camera.east) -| (rewardMerge);
\draw[rewardarrow] (rewardMerge) -- (reward.west);
\draw[updatearrow] (reward) -- (update);

\end{tikzpicture}%
}
\caption{RL training overview. The model first produces long-horizon autoregressive rollouts, then samples short clips for video-quality and camera-control reward-model evaluation. The fused reward drives a moderated DiffusionNFT soft update, decoupling rollout horizon from the optimization window.}
\label{fig:rl-alignment}
\end{figure}

DMD distillation enables efficient autoregressive generation,
but reducing the number of denoising steps can \emph{degrade video generation quality}
and \emph{weaken camera controllability}.
After DMD distillation,
we further train the model with reinforcement learning (RL) \citep{schulman2017proximal,chu2026gpg}
as a post-training stage to enhance video quality and strengthen camera following.

RL after DMD distillation needs to be applied carefully.
Because the model already runs with very few denoising steps,
strong reward updates can make RL training highly unstable.
To keep training stable, we adopt a gradual update strategy,
allowing the model to change step by step.
With this update schedule, RL training proceeds stably
and avoids early collapse.

For each text-image-camera condition,
the current model generates several long-horizon rollout candidates.
The full rollout preserves the autoregressive context,
and short clips sampled from candidates are used for reward computation and DiffusionNFT training \citep{zheng2025diffusionnft,zhang2026astrolabe}
(Figure~\ref{fig:rl-alignment}).
Because the reward models operate on short temporal windows,
the RL update backpropagates through sampled clips rather than the full rollout,
keeping GPU memory within a practical range.
We adopt two reward models:
one measures horizontal translation and rotation accuracy,
and the other evaluates the visual quality of generated clips.
KL regularization balances the two rewards
and keeps the updated model close to the original DMD-distilled model.

\FloatBarrier

RL post-training makes the model better at following camera commands
and improves the visual quality of the generated videos.
At the same time,
the model retains the key strengths of DMD distillation:
long-horizon video generation remains stable,
and the few-step inference efficiency is not compromised.

\section{Inference Acceleration}
For interactive deployment, our inference pipeline enables streaming autoregressive generation while further accelerating DiT denoising and VAE decoding to reduce latency.

Our deployment target is interactive streaming generation with bounded latency. We first define the shared chunk-wise autoregressive inference interface for camera-controlled T2V and I2V, then summarize the optimizations that make it run in real time. In our asynchronous deployment, 8 RTX 5090 GPUs jointly execute DiT denoising and VAE decoding, reaching up to 16 FPS.

\subsection{Autoregressive Streaming Inference}

\begin{figure}[H]
    \centering
    \includegraphics[width=0.8\linewidth]{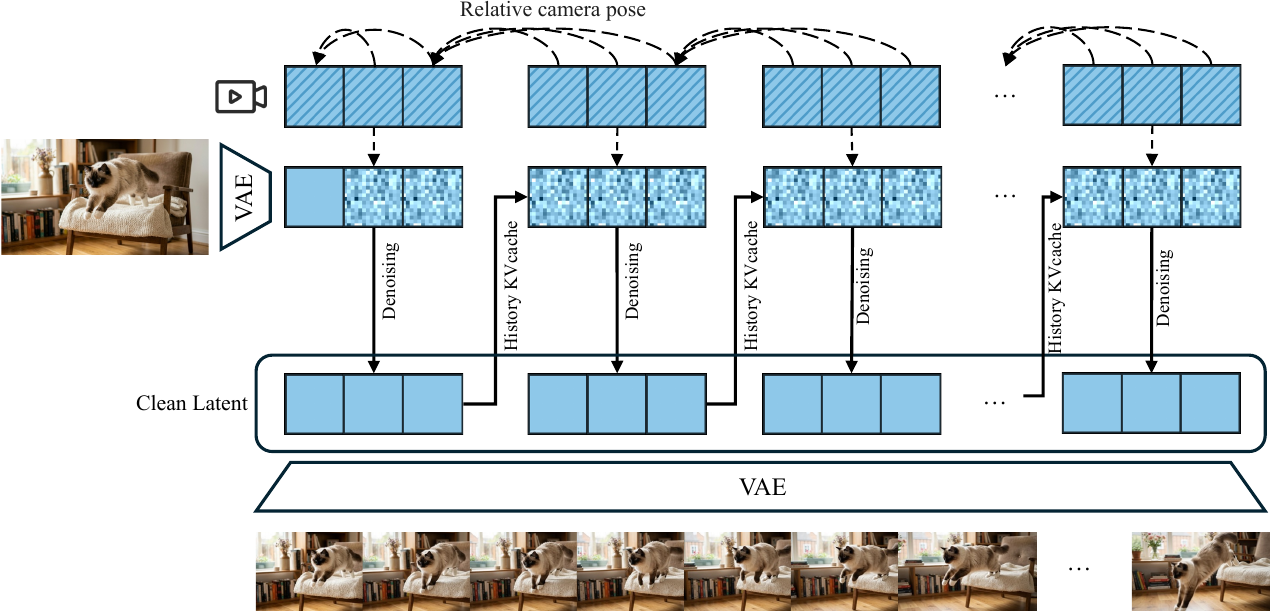}
    \caption{Autoregressive streaming inference. The distilled sampler generates chunks from noise, updates the rolling KV cache, and uses chunk-relative camera controls.}
    \label{fig:autoregressive-streaming-inference}
\end{figure}

As shown in Figure~\ref{fig:autoregressive-streaming-inference}, the video is generated chunk by chunk: each chunk starts from noise, is denoised by the distilled few-step sampler under the text prompt, the chunk-relative camera trajectory, and the rolling KV cache, and then writes its generated tokens back into the cache for subsequent chunks. This keeps inference streaming-friendly because the model only needs to carry a autoregressive history rather than regenerate previous video content.

For camera-controlled I2V, the inference procedure is almost identical to T2V. The only difference is the first chunk: its first frame is replaced by the input image, which anchors the generated video to the reference frame. All later chunks follow the same rolling-cache procedure as T2V and continue from the accumulated generated history.

Camera controls are represented in a chunk-relative form, following E-PRoPE relative camera conditioning \citep{li2025cameras}. The first chunk uses poses relative to its first frame, while each later chunk uses poses relative to the last frame of the previous chunk. This chunk-local relative parameterization keeps the camera condition aligned with the current autoregressive context and prevents the conditioning signal from weakening over long sequences.

\subsection{Inference Optimizations}

For DiT denoising, we combine precision, parallelism, kernel, and timestep-level optimizations. The attention layers use INT8 SageAttention \citep{zhang2025sageattention}, while FFN layers are quantized to FP8 with AngelSlim \citep{cen2026angelslim}. Long spatio-temporal token sequences are sharded across GPUs with sequence parallelism, synchronizing only the required attention and normalization statistics to reduce per-GPU activation memory while preserving full-sequence computation. We further rewrite frequent Transformer-block operators as fused Triton kernels, combining elementwise operations, layout transforms, and small reductions to reduce intermediate allocations and kernel-launch overhead. Following TeaCache \citep{liu2025teacache}, we also reuse denoising residuals in empirically stable timestep regions, skipping selected Transformer-block forward passes when adjacent-step residuals change only marginally.

For VAE decoding, we adopt the Matrix-Game 3.0 VAE \citep{wang2026matrixgame} as the VAE decoder with a 75\% pruning ratio, reducing single-chunk decoding to approximately 0.25 seconds. After the first iteration, \texttt{torch.compile} further reduces later decoding latency. We also follow ParaVAE \citep{riseai2026paravae}: the latent video is split mainly along height, each GPU decodes a local patch, and the decoded patches are gathered into the final video, reducing peak per-GPU memory.

For serving, we use asynchronous pipeline parallelism to overlap the VAE decoding of chunk $k$ with the control reception, KV-cache update, and DiT denoising of chunk $k+1$. This hides most VAE latency behind diffusion computation and enables continuous decoded-chunk emission for real-time interaction.

\section{Evaluation}
\newcommand{\evalmodel}{DreamX-World-1.0-5B}

\begin{figure}[H]
\begin{center}
\includegraphics[width=\textwidth]{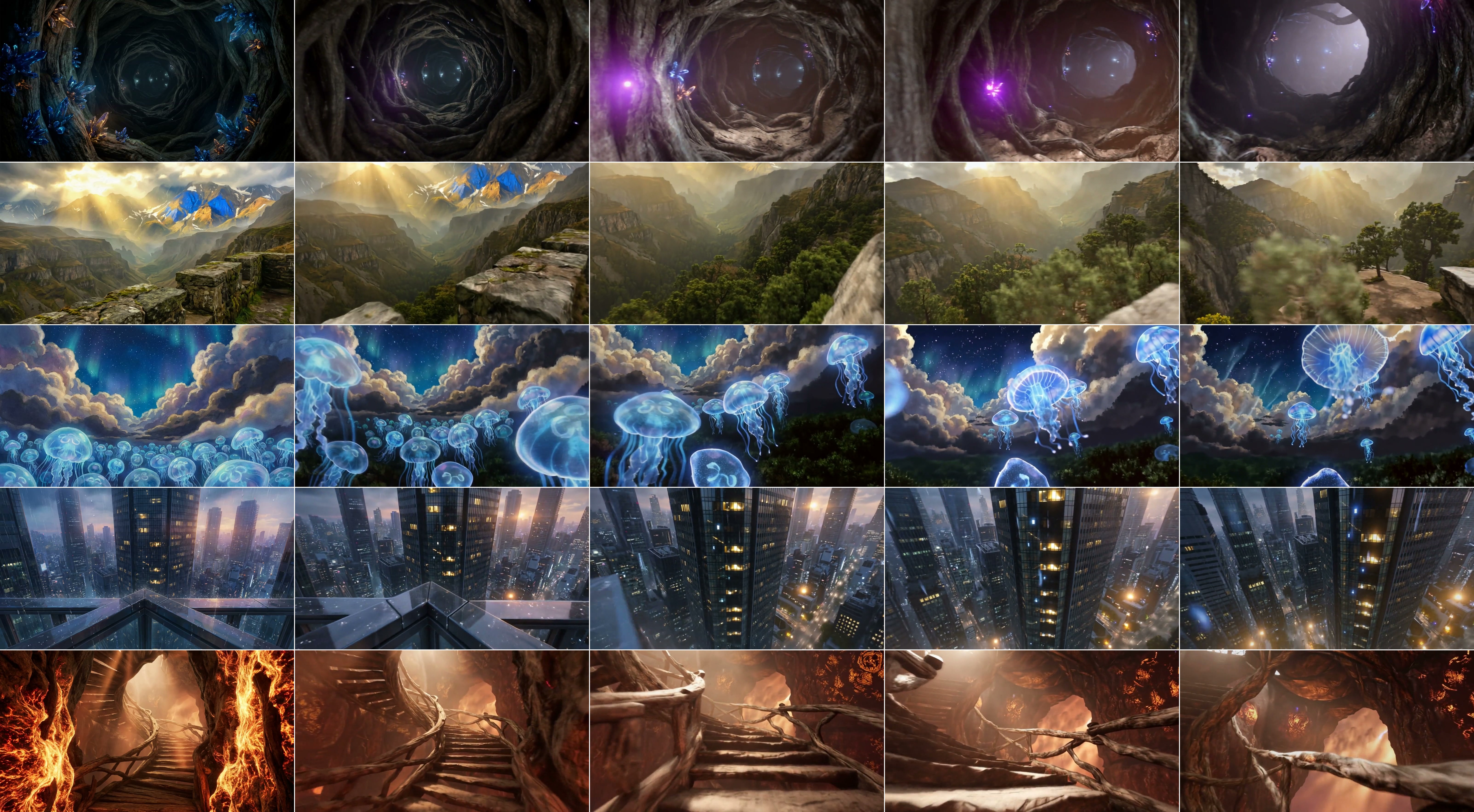}
\end{center}
\caption{Qualitative results of \evalmodel{}. Each row shows five uniformly sampled keyframes from a generated video sequence under different scene types and camera controls.}
\label{fig:qual-base}
\end{figure}


Since the evaluation of interactive world models is still at an exploratory stage, we devise our own evaluation suite that jointly probes camera controllability, perceptual quality, long-horizon behavior, and memory consistency. We first inspect the model qualitatively. Fig.~\ref{fig:qual-base} illustrates generated sequences across diverse scene types, camera trajectories, and visual styles. As the camera moves through the scene, the model produces smooth, temporally coherent transitions while maintaining high visual fidelity across heterogeneous scenes and styles. Additional samples and interactive demos are available on our project page\footnote{\url{https://amap-ml.github.io/DreamX_World}}. We then turn to the quantitative evaluation detailed below.

We compare \evalmodel{} against two representative open-source world models---HY-WorldPlay 1.5 \citep{hyworld2025} (8B) and LingBot-World \citep{gao2026lingbotworld} (14B)---through quantitative evaluation on three complementary axes: basic evaluation on 5-second clips covering camera controllability and visual quality (\cref{tab:basic-metrics}), long-horizon evaluation on ${\sim}$30-second rollouts (\cref{tab:long-horizon}), and memory consistency via revisit-based protocols (\cref{sec:memory-eval}). We additionally conduct a blind human preference study (\cref{sec:user-study}).

\subsection{Basic Evaluation}

We begin with a basic evaluation on standard 5-second generated videos, the most commonly used inference duration, focusing on camera controllability and visual quality. Extended evaluation over longer horizons is reported in \cref{tab:long-horizon}.

\paragraph{\bf{Camera Control Metric.}}
We follow the camera controllability evaluation in WorldScore \citep{duan2025worldscore}, but replace its pose estimator with MegaSaM \citep{li2025megasam} for more accurate camera pose recovery. Beyond adopting the evaluation protocol, we further enrich the evaluation camera trajectories and test data. On the trajectory side, we augment the original set with commonly encountered real-world camera motions such as upward tilt, downward tilt, and diagonal movement, which are under-represented in existing benchmarks yet frequent in practical interactive scenarios. On the data side, our evaluation set covers a broad spectrum of visual domains including AI-generated imagery, stylized content, and simulation-rendered scenes, ensuring that controllability is assessed across a wide range of appearance distributions. The camera control error is computed as:
\begin{equation}
    e_{\mathrm{camera}} = \sqrt{e_{\theta} \cdot e_{t}},
\end{equation}
where $e_{\theta}$ and $e_{t}$ are scale-invariant rotation and translation errors with respect to the ground-truth trajectory, respectively. 
We compute the camera errors across all frames of the generated video V, which are subsequently normalized to yield a final score, where a higher score denotes superior camera controllability.

\paragraph{\bf{Visual Quality metrics.}}
Following Omni-WorldBench \citep{wu2026omniworldbench}, we assess visual quality from multiple complementary dimensions: imaging quality, temporal flickering, motion smoothness, dynamic degree, and transition detection. Together, these metrics comprehensively evaluate the generated videos in terms of frame-level clarity, inter-frame stability, motion intensity, and temporal coherence.

\paragraph{\bf{Artifact Detection Metric.}}

Complementing the visual quality metrics above, we further introduce a multimodal large language model (Gemini-3.1-Pro) for artifact detection, leveraging its strong perceptual reasoning to flag visible defects in the generated frames. This metric focuses on critical defects and failures during the generation process, such as duplicated limbs, objects instantaneously vanishing, and geometric pass-through, among others. We sample frames from each generated video at 2 FPS and prompt the VLM to output a binary pass/fail judgment per sampled frame. Each test case is evaluated twice and averaged, and the final artifact score is the mean pass rate across the evaluation set.

\begin{table}[H]
\caption{Quantitative comparisons on basic metrics. All scores are normalized to $[0,100]$ with higher being better.}
\label{tab:basic-metrics}
\centering
\renewcommand{\arraystretch}{1.2}
\footnotesize
\setlength{\tabcolsep}{2.8pt}
\begin{tabularx}{\linewidth}{@{}lc*{8}{>{\centering\arraybackslash}X}@{}}
\toprule
\rowcolor{amapblue}
\textcolor{white}{\textbf{Model}} &
\textcolor{white}{\textbf{Params}} &
\textcolor{white}{\textbf{Camera\,$\uparrow$}} &
\textcolor{white}{\textbf{Quality\,$\uparrow$}} &
\textcolor{white}{\textbf{Trans.\,$\uparrow$}} &
\textcolor{white}{\textbf{Flicker\,$\uparrow$}} &
\textcolor{white}{\textbf{Smooth.\,$\uparrow$}} &
\textcolor{white}{\textbf{Dynamic\,$\uparrow$}} &
\textcolor{white}{\textbf{Artifact\,$\uparrow$}} &
\textcolor{white}{\textbf{Overall\,$\uparrow$}} \\
\midrule
\rowcolor{TableRowAlt}
HY-WorldPlay 1.5 & 8B & 65.12 & 68.23 & 98.33 & 96.45 & 99.05 & 66.67 & 71.66 & 80.79 \\
LingBot-World & 14B & 71.73 & 67.76 & 85.00 & 94.94 & 97.06 & 88.33 & 58.33 & 80.45 \\
\midrule
\rowcolor{TableRowAlt}
\textbf{\evalmodel{}} & 5B & 73.75 & 66.75 & 98.33 & 96.17 & 98.79 & 85.83 & 73.75 & \textbf{84.76} \\
\bottomrule
\end{tabularx}
\end{table}

The overall score is the average of all individual metrics. As shown in Tab.~\ref{tab:basic-metrics}, our method achieves the highest camera control score and the best overall score while maintaining competitive visual quality. The combination of \eprope{} camera conditioning and RL-based alignment enables precise trajectory adherence while maintaining competitive perceptual quality. In terms of motion naturalness, our model generates richer and more physically plausible dynamics, benefiting from the diverse camera coverage in the UE data engine and the forcing-based training that encourages robust temporal evolution.

\subsection{Long-Horizon Evaluation}


We further extend the evaluation to approximately 30-second generated rollouts to measure how each metric behaves under the long-horizon regime. We adopt the same set of metrics as the basic evaluation.

\begin{table}[H]
\caption{Long-horizon evaluation on 30s generation rollouts. All scores are normalized to $[0,100]$ with higher being better.}
\label{tab:long-horizon}
\centering
\renewcommand{\arraystretch}{1.2}
\footnotesize
\setlength{\tabcolsep}{2.8pt}
\begin{tabularx}{\linewidth}{@{}lc*{8}{>{\centering\arraybackslash}X}@{}}
\toprule
\rowcolor{amapblue}
\textcolor{white}{\textbf{Model}} &
\textcolor{white}{\textbf{Params}} &
\textcolor{white}{\textbf{Camera\,$\uparrow$}} &
\textcolor{white}{\textbf{Quality\,$\uparrow$}} &
\textcolor{white}{\textbf{Trans.\,$\uparrow$}} &
\textcolor{white}{\textbf{Flicker\,$\uparrow$}} &
\textcolor{white}{\textbf{Smooth.\,$\uparrow$}} &
\textcolor{white}{\textbf{Dynamic\,$\uparrow$}} &
\textcolor{white}{\textbf{Artifact\,$\uparrow$}} &
\textcolor{white}{\textbf{Overall\,$\uparrow$}} \\
\midrule
\rowcolor{TableRowAlt}
HY-WorldPlay 1.5 & 8B & 65.86 & 63.02 & 91.00 & 97.00 & 99.11 & 52.00 & 14.00 & 68.85 \\
LingBot-World & 14B & 63.76 & 60.81 & 54.00 & 96.59 & 97.86 & 87.00 & 12.00 & 67.43 \\
\midrule
\rowcolor{TableRowAlt}
\textbf{\evalmodel{}} & 5B & 62.03 & 64.11 & 80.00 & 96.35 & 98.41 & 75.00 & 17.00 & \textbf{70.41} \\
\bottomrule
\end{tabularx}
\end{table}

As shown in \cref{tab:long-horizon}, \evalmodel{} achieves the highest overall score (70.41), outperforming both HY-WorldPlay 1.5 (68.85, 8B) and LingBot-World (67.43, 14B). Our model attains the best imaging quality and artifact detection scores, demonstrating that the forcing-based architecture sustains higher visual fidelity over long horizons than larger competing models.

\subsection{Memory Evaluation via Revisit Consistency}
\label{sec:memory-eval}

Existing world-model benchmarks such as WorldScore~\citep{duan2025worldscore}
and Omni-WorldBench~\citep{wu2026omniworldbench}
primarily assess short-term properties---visual quality, temporal flickering,
and camera controllability---without requiring the agent
to return to a previously visited region.
Distributional metrics such as FVD and FID capture overall generation quality
but cannot reveal whether a model remembers
the \emph{specific} scene it generated moments ago.
In interactive settings, however, users inevitably revisit earlier locations,
making long-horizon spatial memory a critical yet largely unevaluated capability.

To address this gap, we evaluate memory through \emph{revisit consistency},
which detects frame pairs visiting the same spatial location
at different times and checks whether the generated observations agree.
A practical challenge is that imprecise camera control
can introduce slight viewpoint offsets between revisit pairs,
producing apparent inconsistencies unrelated to memory.
We therefore adopt a multi-level metric suite
so that each level captures a distinct facet of memory
while offering progressively greater robustness to such offsets:
pixel-level fidelity measures exact appearance preservation,
perceptual consistency reflects human-perceived similarity,
semantic identity captures high-level scene content,
place recognition identifies the same location across viewpoint changes,
and geometric structure verifies local feature correspondence.
A temporal-smoothness metric further ensures
that memory is not achieved at the cost of incoherent transitions.

\begin{figure}[H]
    \centering
    \includegraphics[width=0.8\linewidth]{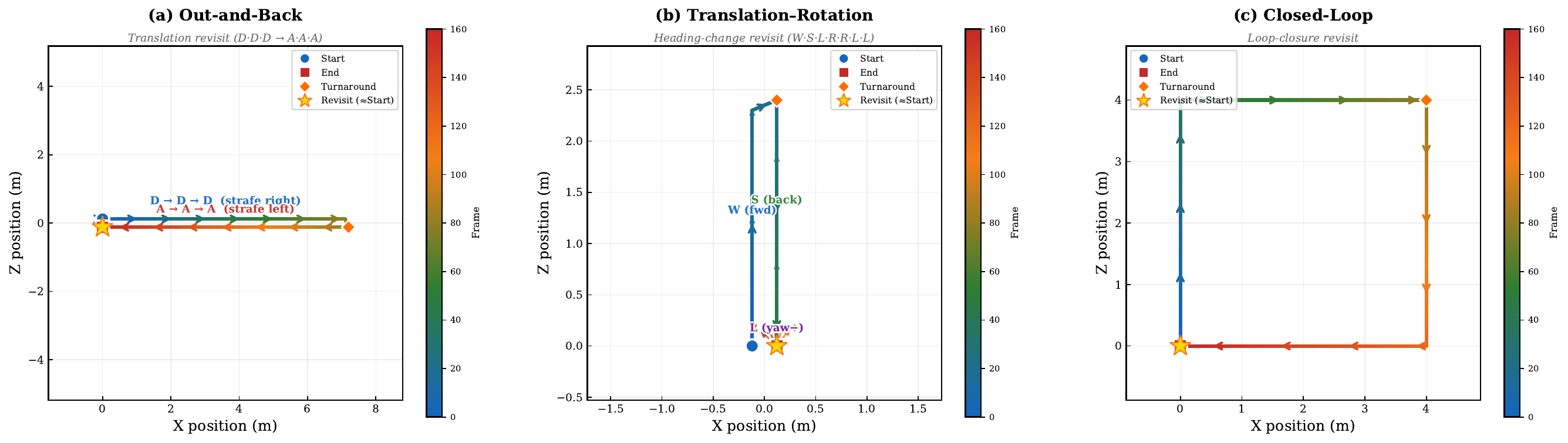}
    \caption{%
        Bird's-eye view of the three evaluation trajectory templates.
        Color encodes temporal progression from start
        (\textcolor{blue}{blue}) to end (\textcolor{red}{red});
        arrows indicate camera movement direction.
        \textbf{(a)}~\emph{Out-and-back}:
        translates laterally (D$\times$3) then reverses (A$\times$3),
        returning with identical orientation.
        \textbf{(b)}~\emph{Translation--rotation}:
        combines translation with heading changes
        (W$\cdot$S$\cdot$L$\cdot$R$\cdot$R$\cdot$L$\cdot$L),
        revisiting from a different yaw angle.
        \textbf{(c)}~\emph{Closed-loop}:
        traverses a rectangular path back to the exact starting pose.
        The $\star$ marker denotes the revisit point
        that coincides with the start.%
    }
    \label{fig:trajectory-templates}
\end{figure}

\paragraph{\bf{Trajectory Construction.}}
We construct camera trajectories that explicitly induce revisits
using simple navigation primitives
(forward/backward translation, left/right rotation).
As shown in Fig.~\ref{fig:trajectory-templates},
three complementary templates are used:
(1)~an \emph{out-and-back} path that revisits with nearly identical orientation,
testing appearance stability;
(2)~a \emph{closed-loop} path that returns to the starting pose,
testing globally consistent layout under loop closure;
and (3)~a \emph{translation-rotation} path that introduces heading changes,
testing place-identity preservation under viewpoint variation.

\paragraph{\bf{Revisit Pair Detection.}}
From camera extrinsics, we extract position
$\mathbf{t}=(t_x,t_y,t_z)$ and yaw $\theta$ for each frame.
A revisit pair $(i,j)$ requires
\begin{equation}
    |\theta_i-\theta_j| \leq \tau_\theta,
    \qquad
    \|\mathbf{t}_i-\mathbf{t}_j\|_2 \leq \tau_t,
\end{equation}
with $\tau_\theta=2^\circ$, $\tau_t=0.1$,
and a minimum temporal gap
$|j-i| \geq \lfloor 0.2\,T \rfloor$
to focus on long-horizon memory.
Among multiple candidates, we select the pair
with the smallest weighted pose distance
$|\theta_i-\theta_j| + 10\,\|\mathbf{t}_i-\mathbf{t}_j\|_2$.

\paragraph{\bf{Metrics.}}
For each revisit pair $(I_i, I_j)$ we compute metrics
spanning all five aspects.
\textbf{(i)~Pixel-level fidelity.}
\textbf{PSNR} and \textbf{SSIM} measure signal-to-noise ratio
and local structural similarity, respectively,
providing strict low-level references sensitive to any scene-content drift.
\textbf{(ii)~Perceptual consistency.}
\textbf{LPIPS}~\citep{zhang2018lpips} measures perceptual distance
in a deep feature space,
capturing whether the two frames \emph{look} different to a human observer;
lower values indicate higher consistency.
\textbf{(iii)~Semantic identity.}
\textbf{DINO-Sim} computes cosine similarity
between frozen DINOv2~\citep{oquab2023dinov2} features of the two frames;
a drop signals altered semantic content upon revisitation.
\textbf{(iv)~Place recognition.}
\textbf{VPR-Sim} uses global descriptors
from MutualVPR~\citep{gu2026mutualvpr},
which is trained to retrieve the same location across viewpoint changes,
thereby reducing the confounding effect of camera-control error.
\textbf{(v)~Geometric structure.}
\textbf{SP-Match} detects up to 1024
SuperPoint~\citep{detone2018superpoint} keypoints per frame
and matches them with LightGlue~\citep{lindenberger2023lightglue};
we report the matching ratio
$r_{\text{match}} = N_{\text{match}} / \min(N_i, N_j)$.
We additionally report \textbf{CLIP-Video}~\citep{radford2021clip},
the average CLIP similarity between consecutive frames,
as a complementary temporal-smoothness measure.

\paragraph{\bf{Gain-based Scoring.}}
Absolute similarity scores can be inflated by slow camera movement
rather than genuine memory.
We therefore sample non-revisit baseline pairs
with a matched temporal-gap distribution
and report all metrics as gains:
$S_{\mathrm{revisit}} - S_{\mathrm{baseline}}$ for similarity metrics,
$S_{\mathrm{baseline}} - S_{\mathrm{revisit}}$ for LPIPS,
so that positive values consistently indicate better memory.
CLIP-Video is reported as an absolute value.

\begin{table}[H]
\caption{Memory consistency evaluation on 10-second generated videos.
Metrics span pixel-level fidelity ($\Delta$PSNR, $\Delta$SSIM),
perceptual consistency ($\Delta$LPIPS),
semantic identity ($\Delta$DINO-Sim),
place recognition ($\Delta$VPR-Sim),
and geometric structure ($\Delta$SP-Match).
All gains are over non-revisit baselines; CLIP-V is absolute.
Higher is better for all columns.}
\label{tab:memory-consistency}
\centering
\renewcommand{\arraystretch}{1.1}
\setlength{\tabcolsep}{3.5pt}
\footnotesize
\begin{tabular}{@{}lccccccc@{}}
\toprule
\rowcolor{amapblue}
\textcolor{white}{\textbf{Model}} &
\textcolor{white}{$\Delta$\textbf{PSNR}} &
\textcolor{white}{$\Delta$\textbf{SSIM}} &
\textcolor{white}{$\Delta$\textbf{LPIPS}} &
\textcolor{white}{$\Delta$\textbf{DINO-Sim}} &
\textcolor{white}{$\Delta$\textbf{VPR-Sim}} &
\textcolor{white}{$\Delta$\textbf{SP-Match}} &
\textcolor{white}{\textbf{CLIP-V}} \\
\midrule
\rowcolor{TableRowAlt}
LingBot-World & 0.61 & 0.019 & 0.039 & 0.090 & 0.100 & 0.088 & 0.987 \\
HY-WorldPlay 1.5 & 3.19 & 0.079 & 0.202 & 0.200 & 0.110 & \textbf{0.251} & \textbf{0.992} \\
\midrule
\rowcolor{TableRowAlt}
\textbf{\evalmodel{}} & \textbf{3.92} & \textbf{0.098} & \textbf{0.232} & \textbf{0.246} & \textbf{0.142} & 0.216 & 0.991 \\
\bottomrule
\end{tabular}
\end{table}

As shown in Table~\ref{tab:memory-consistency},
\evalmodel{} achieves the highest gains on pixel-level, perceptual,
semantic, and place-recognition metrics,
demonstrating stronger memory at every level of abstraction.
HY-WorldPlay 1.5 leads on SP-Match and CLIP-Video,
while LingBot-World shows lower gains across all revisit metrics.

\subsection{Human Preference Study}
\label{sec:user-study}

To complement automatic metrics, we conduct a blind side-by-side human
preference study. Each trial compares \evalmodel{} with one baseline under the
same prompt, initial condition, camera/action trajectory, and playback setting.
The model identities are anonymized and the left-right order is randomized.
Assessors report whether \evalmodel{} wins, ties, or loses along four dimensions:
overall preference, camera control, visual quality, and artifact detection. We
report all percentages from the perspective of \evalmodel{}.

\begin{figure}[H]
    \centering
    \includegraphics[width=\linewidth]{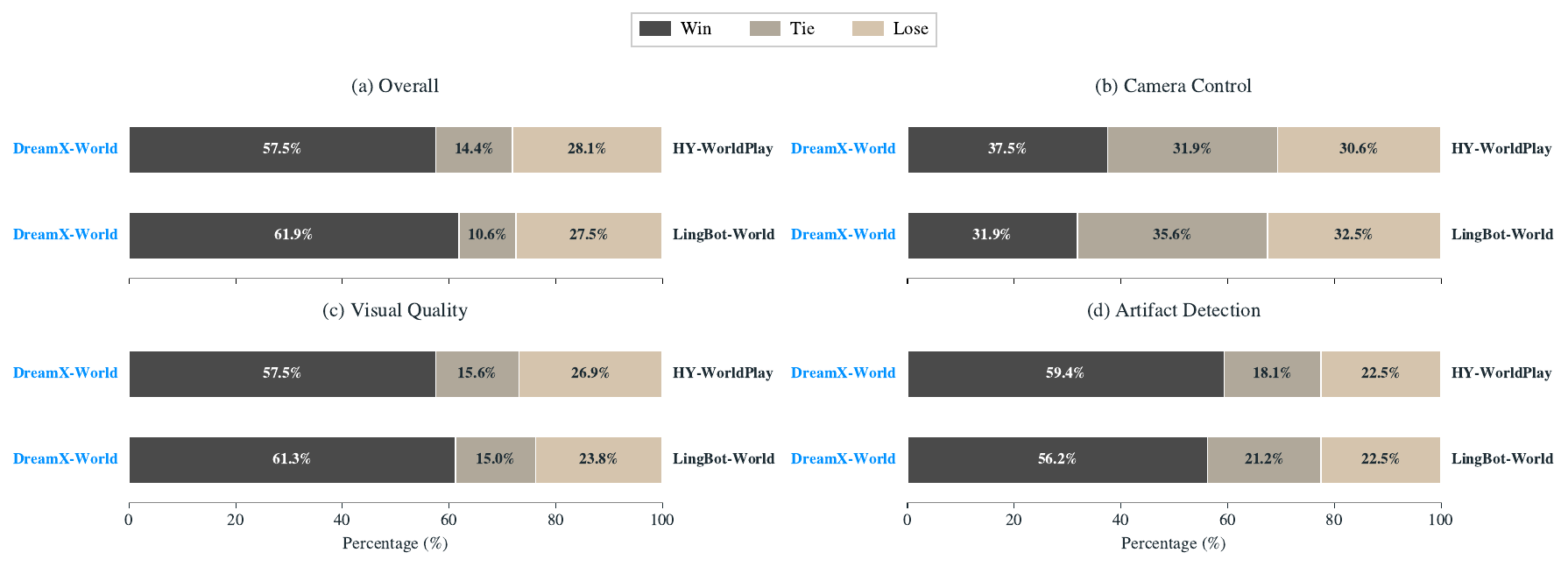}
    \caption{Human preference study comparing \evalmodel{} with HY-WorldPlay 1.5
    and LingBot-World. Each horizontal stacked bar reports Win/Tie/Lose
    percentages from the perspective of \evalmodel{} under blind side-by-side
    comparison. \evalmodel{} is preferred in overall preference, visual quality,
    and artifact detection, while camera-control judgments are close with many
    ties, indicating comparable perceived controllability.}
    \label{fig:user-study}
\end{figure}

As shown in \cref{fig:user-study}, \evalmodel{} is preferred more often in overall preference, achieving win/tie/loss rates of 57.5/14.4/28.1 against HY-WorldPlay 1.5
and 61.9/10.6/27.5 against LingBot-World. The preference is also consistent in
visual quality, where \evalmodel{} obtains 57.5\% and 61.3\% win rates against
the two baselines, and in artifact detection, where it wins 59.4\% and 56.2\%
of comparisons. Camera-control judgments are closer, with higher tie rates,
suggesting comparable perceived controllability among the systems under
side-by-side viewing. Overall, the human study supports the automatic
evaluation: \evalmodel{} improves perceptual quality and artifact robustness while
maintaining competitive camera controllability.

\FloatBarrier

\section{Related Work}

\paragraph{\bf{Video Generation and Interactive World Models.}}
Diffusion models~\citep{ho2020denoising, song2021scorebasedgenerativemodelingstochastic, peebles2023scalable, song2021scorebasedgenerativemodelingstochastic} have become a standard approach to video generation, progressing
from simple image synthesis~\citep{Chu_2025_ICCV, lei2023maskeddiffusionmodelsfast, lei2026vaeendtoendpixelspacegenerative} to complex cinematic video generation~\citep{wan2025wan, feng2026enhancing}.
Current interactive world models extend this setting by conditioning future observations
on user inputs or agent actions. Early systems primarily target gaming scenarios.
Genie~\citep{bruce2024genie} learns action-conditioned environments from videos, GameNGen~\citep{valevski2024gamengen} demonstrates real-time neural game simulation, and GameGen-X~\citep{che2025gamegen} supports interactive open-world game
generation. More recent systems move toward
general-domain video worlds with camera navigation, streaming generation, or
object interaction
\citep{hyworld2025,gao2026lingbotworld,hong2025relic,
wang2026matrixgame,zhao2026minwm,gu2026worldcraft}.
Our work follows this general direction and studies camera control,
long-horizon generation, scene memory, event control, real-time interaction, and world model evaluation.

\paragraph{\bf{Camera-controlled Video Generation.}}
Camera conditioning has been introduced through explicit motion features,
camera embeddings, and geometric representations. MotionCtrl separates camera
and object motion control \citep{wang2024motionctrl}, while CameraCtrl injects
camera trajectories into pretrained video diffusion models
\citep{he2025cameractrl}. AC3D analyzes the representation and training choices
required for 3D camera control in video DiTs \citep{bahmani2025ac3d}. PRoPE
instead incorporates camera intrinsics and extrinsics directly into
self-attention as a projective relative positional encoding
\citep{li2025cameras}. Following this, 
we further introduce \eprope{}, which applies camera-aware attention to a spatially reduced token set, significantly improving computation efficiency.

\paragraph{\bf{Long-horizon Generation and Memory.}}
Autoregressive video generation enables streaming but introduces exposure bias
and accumulated errors. Self-Forcing~\citep{huang2025selfforcing} address this by directly training on model outputs, while Stable Video Infinity~\citep{li2025stablevideoinfinity} adds prediction error injection to model input to reduce its reliance on clean frames, mitigating the negative impact of imperfect context during inference. LongLive and Infinity-RoPE primarily study long-context autoregressive generation
\citep{yang2025longlive,yesiltepe2025infinityrope}. Causal Forcing identifies
the mismatch between bidirectional teachers and causal students and proposes distillation from an autoregressive teacher
\citep{zhu2026causalforcing,zhao2026causalforcingplusplus}.
Beyond recent context, memory-based methods retrieve or compress earlier
observations to preserve scene identity during revisits
\citep{yu2025contextmemory,sun2025worldplay,hong2025relic,
yu2026mosaicmem,wang2026matrixgame}. Our training pipeline combines
autoregressive distillation and rollout-based correction with geometry-guided
retrieval and error-recycled memory conditioning.

\paragraph{\bf{Efficient Sampling.}}
Existing popular few-step diffusion reduces sampling cost primarily through distribution matching
distillation \citep{yin2024onestep,yin2024improved}. Interactive video systems
combine such distillation with causal generation and KV caching to produce
frames incrementally
\citep{huang2025selfforcing,yang2025longlive,zhu2026causalforcing}.
Complementary systems techniques reduce the cost of individual model
components~\citep{zhang2025sageattention, wang2026fasa, liu2025teacache, riseai2026paravae}: SageAttention provides low-precision attention
\citep{zhang2025sageattention}, TeaCache reuses intermediate results across
diffusion timesteps \citep{liu2025teacache}, and ParaVAE distributes VAE
decoding across devices \citep{riseai2026paravae}. We combine few-step
autoregressive generation with quantized DiT execution, residual reuse,
parallel VAE decoding, and asynchronous serving.

\paragraph{\bf{Reinforcement Learning.}}
Reinforcement learning enables diffusion models to optimize non-differentiable
objectives such as perceptual quality and prompt alignment. DDPO formulates
denoising as a sequential decision process \citep{black2023ddpo}, while
Flow-GRPO extends online policy optimization to flow-matching models
\citep{liu2025flowgrpo}. DiffusionNFT instead applies negative-aware
fine-tuning directly to the forward process, avoiding reverse-process likelihood
estimation and solver-specific training \citep{zheng2025diffusionnft}.
However, aggressive reward optimization can reduce generative diversity
\citep{chen2025taming}.
For video world models, reinforcement learning must additionally preserve
temporal coherence during autoregressive rollout. WorldCompass combines
clip-level sampling with interaction and visual-quality rewards
\citep{wang2026worldcompass}, while Astrolabe applies forward-process
reinforcement learning to distilled autoregressive video models using streaming
rollouts and local optimization windows \citep{zhang2026astrolabe}.
Our reinforcement learning stage follows this setting: long rollouts provide temporal
context, short clips serve as reward-bearing units, and camera-control and
video-quality rewards guide a conservative update that preserves the distilled
prior.

\paragraph{\bf{World Model Evaluation}} 
World-model evaluation has gradually moved beyond frame-level video quality~\citep{chen2025finger, ling2025vmbench}
toward controllability, consistency, and interactive response. WorldScore~\citep{duan2025worldscore}
evaluates world generation under prescribed camera trajectories along
controllability, quality, and dynamics. Omni-WorldBench\citep{wu2026omniworldbench} focuses on whether user
interactions produce the intended outcomes and intermediate state transitions. WBench~\citep{ying2026wbenchcomprehensivemultiturnbenchmark} further considers multi-turn navigation, subject actions, event editing, and perspective switching, measuring interaction adherence, consistency, and physical compliance. Complementary to these benchmarks, our evaluation separately examines short-video camera control and visual quality, long-horizon autoregressive generation, and revisit consistency over extended trajectories.

\section{Limitations}

Although \method{} improves interaction, controllability, and efficiency,
several challenges remain.
First, long-horizon visual and geometric consistency is still difficult:
generated worlds may drift drastically in object appearance or layout
after extended interaction.
Second, control signals like caption, camera and event may conflict
when an event produces visual content that is incompatible with the future observations in certain world settings specified by the caption.
Third, automatic evaluation of world models remains imperfect;
benchmarks such as Omni-WorldBench V2 are important,
but human and task-based evaluation will remain necessary for open-ended interaction.

\section{Conclusion}

We presented \method{},
a general-purpose interactive world model.
The central lesson is that world modeling is a full-stack problem: data curation, training, evaluation, and inference acceleration must be organized and improved from a global perspective. Driven by this full-stack perspective, \method{} establishes a promising foundation for the next generation of interactive world models.

\paragraph{Future work.}
Two directions are especially promising.
First, character-centric world models.
It primarily focuses on maintaining persistent character
identity, coordinating character actions with freely moving cameras, and supporting
richer multi-character interactions over long horizons.
Second, native audio-visual world models. It jointly generates synchronized
speech, ambient sound, and action-dependent audio, while also utilizing sound as an
interactive signal for events and scene dynamics.
Together with stronger memory and physical reasoning, these extensions would
move world models toward more embodied, expressive, and immersive simulation.

\section*{Authors}

\paragraph{Team Members.}
\emph{Team members are listed alphabetically by last name (and by first name
where last names are identical). The ordering does not indicate relative
contributions.}

Yancheng Bai, Rui Chen, Xiangxiang Chu, Rujing Dang, Hao Dou, Bingjie Gao, Qiwen Gu, Siyu Hong,
Jiachen Lei, Geng Li, Jifan Li, Ruimin Lin, Qingfeng Shi, Bingze Song,
Lei Sun, Jing Tang, Ruitian Tian, Jun Wang, Jiahong Wu, Pengfei Zhang,
Shen Zhang, and Jiashu Zhu.

\paragraph{Acknowledgements.}
We thank the AMAP-ML team for computational infrastructure, engineering support,
and discussions that made this work possible.

\bibliographystyle{plainnat}
\bibliography{iclr2026_conference}

\end{document}